\documentclass[runningheads]{llncs}

\usepackage{iciap}

\usepackage{iciapabbrv}

\usepackage{graphicx}
\usepackage{booktabs}

\usepackage[accsupp]{axessibility}  %

\usepackage{hyperref}

\usepackage{orcidlink}
\usepackage[table]{xcolor}
\definecolor{lightblue}{RGB}{220,230,250}

\usepackage{caption}

\begin{document}

\title{DOLFIN: Balancing Stability and Plasticity in Federated Continual Learning}

\titlerunning{DOLFIN: Balancing Stability and Plasticity in Federated Continual Learning}

\author{Omayma Moussadek\orcidlink{0009-0002-4714-3773} \and
Riccardo Salami\orcidlink{0009-0002-0704-5810} \and
Simone Calderara\orcidlink{0000-0001-9056-1538}}

\authorrunning{O. Moussadek, R. Salami, S. Calderara}
\institute{
AImageLab, University of Modena and Reggio Emilia, Modena, Italy \\
\email{253399@studenti.unimore.it}, 
\email{riccardo.salami@unimore.it}, 
\email{simone.calderara@unimore.it}
}

\maketitle

\newcommand{\best}[1]{\textbf{#1}}
\newcommand{\second}[1]{\underline{#1}}
\newcommand{\methname}{\textit{DOLFIN}\xspace}

\begin{abstract}
\textit{Federated continual learning (FCL) enables models to learn new tasks across multiple distributed clients, protecting privacy and without forgetting previously acquired knowledge. However, current methods face challenges balancing performance, privacy preservation, and communication efficiency. We introduce a Distributed Online LoRA for Federated INcremental learning method \textbf{\methname}, a novel approach combining Vision Transformers with low-rank adapters designed to efficiently and stably learn new tasks in federated environments. Our method leverages LoRA for minimal communication overhead and incorporates Dual Gradient Projection Memory (DualGPM) to prevent forgetting. Evaluated on CIFAR-100, ImageNet-R, ImageNet-A, and CUB-200 under two Dirichlet heterogeneity settings, \textbf{\methname} consistently surpasses six strong baselines in final average accuracy while matching their memory footprint. Orthogonal low-rank adapters offer an effective and scalable solution for privacy-preserving continual learning in federated settings.}
\end{abstract}

\section{Introduction}
\label{sec:intro}

Deep learning models increasingly face two interconnected challenges in real-world scenarios: they must learn from data that arrive sequentially while preserving past knowledge and operating under privacy constraints that enforce a decentralized data distribution. \emph{Continual Learning} (CL) and \emph{Federated Learning} (FL) individually tackle these issues, and their combination gives rise to \emph{Federated Continual Learning} (FCL), where both constraints must be satisfied concurrently.
In CL, the main challenge is \emph{catastrophic forgetting}~\cite{mccloskey1989catastrophic}: once data from earlier tasks are no longer available, gradient updates for new tasks can overwrite parameters that encode prior knowledge. Overcoming this problem demands a careful balance between two competing objectives: \textit{plasticity}, the model’s ability to learn new tasks effectively, and \textit{stability}, its ability to retain information acquired from previous tasks.
In parallel, FL allows multiple clients to train a shared model while keeping their data on local devices, thus avoiding direct data sharing and helping to protect privacy.
To address these challenges, inspired by \emph{InfLoRA}~\cite{liang2024inflora} and recent studies on modular compositionality~\cite{porrello2024second}, we propose \methname an FCL method based on Vision Transformers (ViT)~\cite{dosovitskiy2020image} where each encoder layer is equipped with \emph{LoRA}~\cite{hu2022lora} modules. 

We validate \methname on diverse benchmarks under varying data heterogeneity, achieving state-of-the-art performance.

\section{Related Work}
\label{sec:related}

\paragraph{Federated Learning.}

The classical \emph{FL} loop aggregates locally–trained models through weighted parameter averaging (\textsc{FedAvg}) \cite{mcmahan2017communication}.
To handle statistical and system heterogeneity, several variants constrain the local optimisation:
FedProx adds a proximal term to keep client solutions near the server model \cite{li2020federated};
SCAFFOLD expands on this approach and introduces control variates to further regularize local training. \cite{karimireddy2020scaffold};
FedDC lets each client add its estimated drift to its model before upload, so the server aggregates drift-corrected weights \cite{gao2022feddc};
and GradMA overcomes quadratic-programming obstacles by projecting each client’s gradient into a compact memory subspace and redirecting updates so they optimize the local objective while staying close to the server model \cite{luo2023gradma}.
Prototype-based aggregation represents each class with local centroids that are averaged on the server (FedProto) \cite{tan2022fedproto} or calibrated with synthetic IID features \cite{luo2021no}.

\paragraph{Class-Incremental Learning.}

In \emph{CIL} a model meets disjoint label sets over time \cite{van2022three}.
Early methods rely on weight regularization (EWC \cite{kirkpatrick2017overcoming}, SI \cite{zenke2017continual}) or on distillation against previous predictions (LwF) \cite{li2017learning}.
Rehearsal stores real or synthetic samples to replay past tasks, e.g.\ tiny episodic memories \cite{chaudhry2019continual}, dark experience replay \cite{buzzega2020dark}, or iCaRL \cite{rebuffi2017icarl}.
With the advent of large self-attentive backbones \cite{dosovitskiy2020image}, buffer-free \emph{Parameter-Efficient Fine-Tuning} (PEFT) has become prevalent:
L2P \cite{wang2022learning}, DualPrompt \cite{wang2022dualprompt} and CoDA-Prompt \cite{smith2023coda} attach prompt pools that grow with tasks, while CLIP-GLR combines CLIP features with generative replay \cite{frascaroli2024clip}.

\paragraph{Federated CIL} (FCIL) combines the above two settings.
FedWeIT \cite{yoon2021federated} splits client parameters into generic and task-specific subsets via sparse masks.
GLFC \cite{dong2022federated} and its extension LGA \cite{dong2023no} couple local buffers with class-aware gradient compensation; TARGET \cite{zhang2023target} relies on a shared generator to supply rehearsal samples.
Recent FCIL work exploits PEFT:
Fed-CPrompt injects divergence-regularised prompts \cite{bagwe2023fed};
PILoRA integrates LoRA branches guided by aggregated prototypes at the transformer level \cite{guo2024pilora} and Hierarchical Generative Prototypes (HGP) balance the global classifier via hierarchical GMM sampling \cite{salami2024federated}, while LoRM~\cite{salami2024closed} merges client-specific and task-specific LoRA adapters in a closed-form to align them on the global model. 
Our method is a PEFT approach, concentrating on low-rank adapters that are explicitly designed to be interference-free, thereby preserving PEFT’s communication efficiency while eliminating the need for rehearsal and prototype storage.

\section{Methodology}
\label{sec:method}

\begin{figure}[ht]
    \centering
    \includegraphics[width=0.94\textwidth]{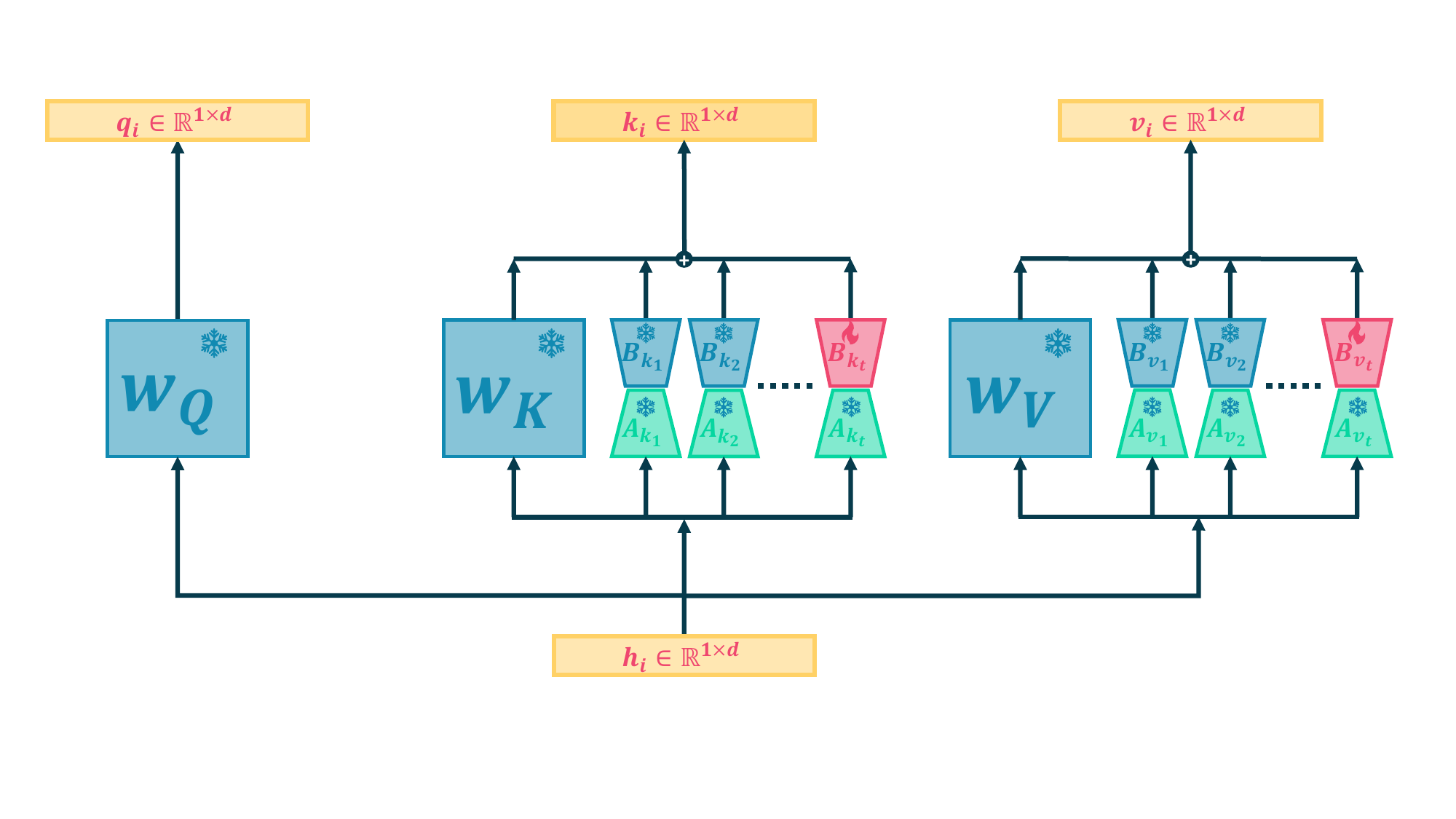}
    \caption[LoRA modules]{At each task, a new matrix \( A_t \) is designed and kept frozen. Only the corresponding matrix \( B_t \) is updated during training. All previous parameters, including earlier \( B\) matrices and pre-trained weights, remain frozen.}
    \label{fig:inflora_modules}
\end{figure}

In our FCL setting, each of the $K$ clients holds a subset of the training data for task $t$, and all clients follow the same task sequence $\{\mathcal{D}_t^k\}_{t=1}^{T}$.
\methname builds on a ViT backbone, where each encoder block is augmented with task-specific LoRA modules on the key and value projections, reparameterizing weights as: 

\begin{equation}
\label{eq:lora}
K = W_{K_{t-1}}x + A_{K_t} B_{K_t} x, \quad
V = W_{V_{t-1}}x + A_{V_t} B_{V_t} x,
\end{equation}

where the input token embedding is $x\in \mathbb{R}^{d}$; the low-rank matrices satisfy $A_{\{K,V\}} \in \mathbb{R}^{d \times r}$ and $B_{\{K,V\}} \in \mathbb{R}^{r \times d}$, with $r \ll d$, while $W_{K_{t-1}}=W_K+\sum_{i=1}^{t-1} A_{K_i}B_{K_i}$ and $W_{V_{t-1}}=W_V+\sum_{i=1}^{t-1} A_{V_i}B_{V_i}$. The two LoRA matrices serve different purposes, to balance plasticity and stability: $B$ matrices are the trainable ones, while $A$ matrices are frozen and their columns span a task-specific update subspace. \Cref{fig:inflora_modules} illustrates how each $A_t$ remains fixed while only the corresponding $B_t$ is learned.

Ideally, to mitigate catastrophic forgetting, each adapter \(A_t\) should project the updates from \(B_t\) into a space that is orthogonal to that of previous tasks, thereby minimizing interference.
Orthogonal adapters \(A_t\) follow a similar principle to spectral re-basin methods~\cite{rinaldi2025update}. 
To enforce this, we constrain the LoRA updates to lie in a subspace orthogonal to the gradient subspace of earlier tasks. 
Since past data is unavailable, we integrate Dual Gradient Projection Memory (DualGPM)~\cite{liang2023adaptive} to maintain an orthonormal basis $\mathcal{M}_t$ that approximates past gradients. The update must satisfy:
\begin{equation}
\operatorname{span}(A_t) \subseteq \mathcal{N}_t \cap \mathcal{M}_t^{\perp},
\label{eq:dual_constraint}
\end{equation}
where $\mathcal{N}_t$ is the gradient subspace of the current task. To enforce this, we project the frozen hidden activations $H_t \in \mathbb{R}^{d \times n}$ onto $\mathcal{M}_t^{\perp}$: $\hat{H}_t = (I - \mathcal{M}_t \mathcal{M}_t^\top) H_t$. Then, SVD on $\hat{H}_t^\top$ provides the top-$r$ singular components. After training, DualGPM updates $\mathcal{M}_t$ by removing from $\mathcal{M}_t^{\perp}$ components aligned with the new task gradients, ensuring continual capacity expansion without overlap.

However, in a federated setting, computing the optimal $A_{t+1}$ with respect to previous tasks is challenging, as the data is distributed across clients. This makes it infeasible to compute $\mathcal{M}_t$ on the complete set $\{\mathcal{D}_t\}_t$. 
To address this, at each task, the server initially broadcasts matrices $A_t$ and $B_t$ to clients, which independently train their local matrices $B_t^k$, keeping other modules fixed. These local updates are then aggregated via weighted averaging as $B_t = \sum_k \frac{n_k}{\sum_j n_j} B_t^k$, where $n_k$ reflects the size of client $k$'s dataset and $\sum_{j=1}^{K} n_j$ denotes the total number of samples used in the round. This process integrates knowledge from the current task. Subsequently, each client computes $A_{t+1}^k$ using DualGPM, ensuring orthogonality to previous tasks' gradient subspaces:

\begin{equation}
\operatorname{span}\bigl(A_{t+1}^{k}\bigr)
\subseteq
\mathcal{N}{t}^{k},\cap,\bigl(\mathcal{M}{t}^{k}\bigr)^{\perp}.
\label{eq:next_A}
\end{equation}

The server averages these local matrices to form the unified $A_{t+1}$, maintaining interference-free continual learning without accessing past data or other clients’ data. 
During inference, the central model is used to classify all seen classes.

\captionsetup{skip=-20pt}

\section{Experiments}
\label{sec:exp}
\paragraph{Datasets and Preprocessing}
We evaluate \methname on four image classification benchmarks commonly adopted in FCL: CIFAR-100~\cite{krizhevsky2009learning}, ImageNet-R~\cite{hendrycks2021many}, ImageNet-A~\cite{hendrycks2021natural}, and CUB-200~\cite{wah2011caltech}. Each dataset is split into 10 incremental tasks, each with the same number of classes.  To simulate realistic non-IID scenarios, data is distributed across 10 clients using a Dirichlet distribution parameterized by $\beta$. Lower values of $\beta$ correspond to higher heterogeneity and thus a more challenging learning environment, characterized by significant data imbalance among clients. Conversely, higher values represent homogeneous and balanced data distributions. We use $\beta \in \{0.5, 0.1\}$ for CIFAR-100 and ImageNet-R, and $\beta =1.0$ for ImageNet-A and CUB-200 to reflect their different characteristics.

Each dataset is preprocessed according to its specific format. CIFAR-100 images are resized from $32 \times 32$ to $224 \times 224$ using bicubic interpolation, followed by random horizontal flipping and normalization. ImageNet-R and CUB-200 images are resized to $224 \times 224$ and augmented with random flipping. For ImageNet-A, we apply random resized cropping (scale range of (0.05,1.0) and aspect ratio range of ($\frac{3}{4}$, $\frac{4}{3}$)), followed by flipping and normalization. At test time, all images are resized to $256 \times 256$, center-cropped to $224 \times 224$, and normalized.

\paragraph{Evaluated approaches}
We compare \methname with six competitive baselines. From CL we consider EWC, LwF, L2P, and CODA-Prompt, which are adapted to the federated scenario using the \textsc{FedAvg} strategy. We evaluate Fisher-Avg, which uses Fisher information to aggregate client models, and PILoRA, a native FCL parameter-efficient approach; the upper bound is a jointly trained centralized model on the full dataset, free of federated or incremental constraints.

\paragraph{Implementation Details}
We employ a ViT-B/16 backbone~\cite{dosovitskiy2020image} pre-trained on ImageNet-21K~\cite{ridnik2021imagenet}, which remains frozen throughout training for both our method and all baselines to provide a fair comparison. 
All models are trained locally for 5 epochs per task using the AdamW optimizer, with learning rates selected from the range $[10^{-5}, 3 \times 10^{-2}]$ and a batch size of 16. 

\paragraph{Hyperparameter Selection}
All baseline methods are tuned using their original hyperparameter settings, while for \methname, a two-phase grid search was conducted to determine optimal learning rates and rank values, first with coupled learning rates for backbone and head, and then with decoupled rates following the SLCA~\cite{zhang2023slca} strategy. A complete summary of the hyperparameters used for each method and dataset is provided in Table~\ref{tab:hyperparams_all}.

\paragraph{Evaluation Metrics.}
We evaluate all methods using the \emph{Final Average Accuracy (FAA)}, a widely adopted metric in the FCL literature. FAA measures the mean classification accuracy across all tasks after the entire incremental training process has concluded. Formally, let $R_{T,i}$ denote the accuracy on task $i$ evaluated after completing the final task $T$. Then, FAA is defined as:
\begin{equation}
\text{FAA} = \frac{1}{T} \sum_{i=1}^{T} R_{T,i}.
\label{eq:FAA}
\end{equation}

\begin{table}[tb]
  \caption [FCL results on CIFAR100, ImageNetR, ImageNetA, CUB200]{Performance on CIFAR-100 and ImageNet-R with $\beta \in \{1.0, 0.5\}$, ImageNet-A and CUB-200 with $\beta = 1.0$. Best results are in bold, second-best underlined.}
  \label{tab:fcl_selected_results}
  \centering
  \small
  \setlength{\tabcolsep}{5pt}
  \begin{tabular}{l|cc|cc|c|c}
    \toprule
    \textbf{Method} &
      \multicolumn{2}{c|}{\textbf{CIFAR-100}} &
      \multicolumn{2}{c|}{\textbf{ImageNet-R}} &
      \textbf{ImageNet-A} &
      \textbf{CUB-200} \\[2pt]
      & $\beta{=}1.0$ & $\beta{=}0.5$
      & $\beta{=}1.0$ & $\beta{=}0.5$
      & $\beta{=}1.0$
      & $\beta{=}1.0$ \\
    \midrule
    Joint           &       & \best{92.75} &       & \best{84.02} & \best{54.64} & \best{86.04} \\
    \midrule
    EWC             & 75.04 & 78.46 & 54.48 & 58.93 & 10.86 & 31.46 \\
    LwF             & 63.68 & 62.87 & 52.55 & 54.03 &  8.89 & 25.25 \\
    Fisher-AVG      & 75.56 & 76.10 & 56.60 & 58.68 & 11.59 & 30.45 \\
    L2P             & \second{85.12} & \second{83.88} & \second{67.90} & 42.08 & \second{20.14} & 56.23 \\
    CODA-Prompt     & 84.91 & 82.25 & 66.23 & \second{61.18} & 18.30 & 42.53 \\
    PILoRA          & 75.75 & 76.48 & 53.53 & 53.67 & 19.62 & \best{61.11} \\
    \midrule
    \rowcolor{lightblue}
    \textbf{\methname} & \best{86.58} & \best{85.27} &
                       \best{74.53} & \best{69.58} &
                       \best{35.75} & \second{60.25} \\
    \bottomrule
  \end{tabular}
  \label{tab:main}
\end{table}

\paragraph{Results.}
Table~\ref{tab:fcl_selected_results} reports FAA across benchmarks. \textit{\methname} consistently outperforms all baselines on CIFAR-100 and ImageNet-R under both Dirichlet settings and achieves a large margin on the challenging ImageNet-A. On CUB-200, it performs comparably to PILoRA. These results confirm \methname's effectiveness in heterogeneous FCL, combining strong generalization with efficient adaptation.

\begin{table}[tb]
\centering
\resizebox{\textwidth}{!}{
\begin{tabular}{lllll}
\toprule
\textbf{Method} & \textbf{CIFAR-100} & \textbf{ImageNet-R} & \textbf{ImageNet-A} & \textbf{CUB-200} \\
\midrule
\textbf{Dirichlet} & $\beta = 0.5,\ 1.0$ & $\beta = 0.5,\ 1.0$ & $\beta = 1.0$ & $\beta = 1.0$ \\
\midrule
EWC         & \textit{lr}: 1e-5 & \textit{lr}: 1e-5 & \textit{lr}: 1e-5 & \textit{lr}: 1e-5 \\
LwF         & \textit{lr}: 1e-5 & \textit{lr}: 1e-5 & \textit{lr}: 1e-5 & \textit{lr}: 1e-5 \\
FisherAVG   & \textit{lr}: 1e-5 & \textit{lr}: 1e-5 & \textit{lr}: 1e-5 & \textit{lr}: 1e-5 \\
L2P         & \textit{lr}: 3e-2 & \textit{lr}: 3e-2 & \textit{lr}: 3e-2 & \textit{lr}: 3e-1 \\
CODA-P      & \textit{lr}: 1e-3 & \textit{lr}: 1e-3 & \textit{lr}: 1e-2 & \textit{lr}: 1e-3 \\
PILoRA      & \textit{lr}: 2e-2 & \textit{lr}: 2e-2; $lr_{\text{pr}}$: 1e-4 & \textit{lr}: 1e-2; $lr_{\text{pr}}$: 1e-4 & \textit{lr}: 1; $lr_{\text{pr}}$: 1e-4 \\
\midrule
\rowcolor{lightblue}
\textbf{\methname} & \textit{lr}: 3e-3; \textit{r}: 2 & \textit{lr}: 1e-3; \textit{r}: 64 & \textit{lr}: 3e-2; \textit{lr\_back}: 3e-3; \textit{r}: 32 & \textit{lr}: 1e-2; \textit{r}: 1 \\
\bottomrule
\end{tabular}
}
\caption[Hyperparameters for each method]{Hyperparameters used for each method across CIFAR-100, ImageNet-R, ImageNet-A, and CUB-200 in the FCL setting.}
\label{tab:hyperparams_all}
\end{table}

\section{Conclusion}
\label{sec:conclusion}

This work tackles the dual challenge of CL on non-IID data while preserving client privacy.
We introduced \methname, a ViT method that combines the \emph{communication efficiency} of LoRA with the stability of orthogonal sub-space updates. By freezing the ViT backbone, training only rank-$r$ matrices $\mathbf{B}_t$, and computing interference-free bases $\mathbf{A}_{t+1}$ via DualGPM, the method removes rehearsal buffers and reduces per-round traffic. 
Across four class-incremental benchmarks with two Dirichlet heterogeneity levels, \methname outperforms six strong baselines, confirming that orthogonal low-rank adapters provide a simple yet powerful way to balance plasticity and stability in realistic federated scenarios.

\bibliographystyle{splncs04}
\bibliography{main}

\end{document}